# Mixture Representations for Inference and Learning in Boltzmann Machines

**Neil D. Lawrence**
Computer Laboratory
New Museums Site, Pembroke Street
Cambridge, CB2 3QG, U.K.
neil.lawrence@cl.cam.ac.uk

**Christopher M. Bishop**
Microsoft Research
St. George House, 1 Guildhall Street
Cambridge, CB2 3NH, U.K.
cmbishop@microsoft.com

**Michael I. Jordan**
Center for Biological and Computational Learning
M.I.T., 79 Amherst Street, E10-243
Cambridge, MA 02139, U.S.A.
jordan@psyche.mit.edu

## Abstract

Boltzmann machines are undirected graphical models with two-state stochastic variables, in which the logarithms of the clique potentials are quadratic functions of the node states. They have been widely studied in the neural computing literature, although their practical applicability has been limited by the difficulty of finding an effective learning algorithm. One well-established approach, known as mean field theory, represents the stochastic distribution using a factorized approximation. However, the corresponding learning algorithm often fails to find a good solution. We conjecture that this is due to the implicit uni-modality of the mean field approximation which is therefore unable to capture multi-modality in the true distribution. In this paper we use variational methods to approximate the stochastic distribution using multi-modal *mixtures* of factorized distributions. We present results for both inference and learning to demonstrate the effectiveness of this approach.

## 1 Introduction

The Boltzmann machine (Ackley *et al.*, 1985) is an undirected graphical model whose nodes correspond to two-state stochastic variables, with a particular choice of clique potentials. Specifically, the joint distribution over all states is given by a *Boltzmann* distribution of the form

$$P(S) = \frac{\exp(-E(S)/T)}{Z} \qquad (1)$$

in which $S = \{s_i\}$ denotes the set of stochastic variables, and $E(S)$ denotes the energy of a particular configuration given by a quadratic function of the states

$$E(S) = -\sum_i \left\{ \sum_{j>i} w_{ij} s_i s_j + w_{i0} s_i \right\}. \qquad (2)$$

Here $w_{ij} = 0$ for nodes which are not neighbours on the graph. Throughout this paper we shall choose $s_i \in \{-1, 1\}$, and we shall also absorb the 'bias' parameters $w_{i0}$ into the 'weight parameters' $w_{ij}$ by introducing an additional variable $s_0 = 1$. The 'temperature' parameter $T$ in (1) is in principle redundant since it can be absorbed into the weights. However, in practice it can play a useful role through 'annealing', as discussed in Section 2, although for the moment we set $T = 1$. The normalization factor $Z^{-1}$ in (1) is called the *partition function* in statistical physics terminology, and is given by marginalizing the numerator over all configurations of states

$$Z = \sum_S \exp(-E(S)). \qquad (3)$$

If there are $L$ variables in the network, the number of configurations of states is $2^L$, and so evaluation of $Z$ may require exponential time (e.g. for fully connected models) and hence is, in the worst case, computationally intractable.

The Boltzmann machine is generally used to learn the probability distribution of a set of variables. We therefore partition the variables into a *visible* set $V = \{v_i\}$ whose values are observed, and a *hidden* set $H = \{h_i\}$ whose values are unobserved. The marginal probability of the observed states is obtained by summing over the hidden variables to give

$$P(V) = \sum_H P(H, V) \qquad (4)$$

which can be viewed as a function of the parameters $\{w_{ij}\}$ in which case it represents a likelihood function. A data set then consists of a set of instantiations of the visible variables $V_1, \ldots, V_N$, where it is assumed that these observations are drawn independently from the same distribution. In this case the log likelihood becomes a sum over patterns

$$\ln P(V) = \sum_{n=1}^{N} \ln \left\{ \sum_{H_n} P(H_n, V_n) \right\}. \qquad (5)$$

Here we are implicitly assuming that it is the same set of variables which are observed in each pattern. The formalism is easily generalized to allow arbitrary combinations of missing and observed variables. From now on we suppress the summations over $n$ to avoid cluttering the notation.

Learning in the Boltzmann machine is achieved by maximizing the log likelihood (5) with respect to the parameters $\{w_{ij}\}$ using gradient methods. Differentiating (5) and using (1) and (2) we obtain

$$\frac{\partial \ln P(V)}{\partial w_{ij}} = \langle s_i s_j \rangle_{\mathrm{C}} - \langle s_i s_j \rangle_{\mathrm{F}} \qquad (6)$$

where $\langle \cdot \rangle_{\mathrm{C}}$ denotes an expectation with respect to the *clamped* distribution $P(H|V)$ while $\langle \cdot \rangle_{\mathrm{F}}$ denotes expectation with respect to the *free* distribution $P(H, V)$ so that, for some arbitrary $G(H, V)$,

$$\langle G(H,V) \rangle_{\mathrm{C}} = \sum_H G(H, V) P(H|V) \qquad (7)$$

$$\langle G(H,V) \rangle_{\mathrm{F}} = \sum_V \sum_H G(H, V) P(H, V). \qquad (8)$$

In the case of the clamped expectation, each $s_i$ in (6) corresponding to a visible variable is set to its observed value.

Evaluation of the expectations in (6) requires summing over exponentially many states, and so is intractable for densely conneted models. The original learning algorithm for Boltzmann machines made use of Gibbs sampling to generate separate samples from the joint and marginal distributions over states, and used these to evaluate the required gradients. A serious limitation of this approach, however, is that the gradient is expressed as the difference between two Monte Carlo estimates and is thus very prone to sampling error. This results in a very slow learning algorithm.

In an attempt to resolve these difficulties, there has been considerable interest in approximating the expectations in (6) using deterministic methods based on mean field theory (Peterson and Anderson, 1987; Hinton, 1989). Although in principle this leads to a relatively fast algorithm, it often fails to find satisfactory solutions for many problems of practical interest. In Section 2 we review the variational framework for approximate inference in graphical models, in which we seek to approximate the true distribution over states with some parametric class of approximating distributions. We show that mean field theory can be derived within this framework by using an approximating distribution which is assumed to be fully factorized. It is this severe approximation which is believed to lie at the heart of the difficulties with mean field theory in Boltzmann machines (Galland, 1993). One of its consequences is that the approximating distribution is constrained to be uni-modal, and is therefore unable to capture multiple modes in the true distribution. As a solution to this problem we introduce *mixtures* of factorized distributions in Section 3, and derive the corresponding algorithms for inference and learning. Experimental results on toy problems, and on a problem involving hand-written digits, are presented in Sections 4 and 5. Conclusions are presented in Section 6.

## 2   Variational Inference

We have seen that, for the probability distribution defined by the Boltzmann machine, standard operations such as normalization, or the evaluation of expectations, involve intractable computations for densely conneted graphs. A general framework for making controlled approximations in such cases is provided by variational methods (Jordan *et al.*, 1998). Consider the conditional distribution $P(H|V)$ of the hidden variables given values for the visible variables. Since it is intractable to work directly with this distribution we consider some family of simpler distributions $Q_{\mathrm{C}}(H|V, \theta)$, where the suffix C denotes 'clamped', governed by a set of parameters $\theta$. We can define the closest approximation within this family to be that which minimizes the Kullback-Leibler (KL) divergence

$$\mathrm{KL}(Q_{\mathrm{C}} \| P) = -\sum_H Q_{\mathrm{C}}(H|V, \theta) \ln \left\{ \frac{P(H|V)}{Q_{\mathrm{C}}(H|V, \theta)} \right\} \qquad (9)$$

with respect to $\theta$. The KL divergence satisfies $\mathrm{KL}(Q\|P) \geq 0$, with equality if and only if $Q = P$. One motivation for this definition is that it corresponds to a lower bound on the log likelihood, since we can write

$$
\begin{aligned}
\ln P(V) &= \ln \sum_H P(H,V) \\
&= \ln \sum_H Q_C(H|V,\theta) \frac{P(H,V)}{Q_C(H|V,\theta)} \\
&\geq \sum_H Q_C(H|V,\theta) \ln \frac{P(H,V)}{Q_C(H|V,\theta)} \\
&\triangleq \mathcal{L} \qquad\qquad (10)
\end{aligned}
$$

where we have used Jensen's inequality. The difference between the left and right hand sides of (10) is given by the KL divergence (9). By maximizing $\mathcal{L}$ with respect to $\theta$ we obtain the highest lower bound achievable with the family of distributions $Q_C(H|V,\theta)$.

The goal in choosing a form for the distribution $Q_C(H|V,\theta)$ is to use a sufficiently rich family of approximating distributions that a good approximation to the true distribution can be found, while still ensuring that the family is sufficiently simple that inference remains tractable.

In the case of the Boltzmann machine, we have to deal with the joint distribution $P(H,V)$ and also with the conditional distributions $P(H_n|V_n)$ for each pattern $n$ in the data set. If we approximate the conditional distributions using variational methods, from (10) we have (and again suppressing the sum over $n$ for convenience)

$$
\mathcal{L} = \mathcal{L}_C - \ln Z \qquad\qquad (11)
$$

where we have defined

$$
\begin{aligned}
\mathcal{L}_C = &-\sum_H Q_C(H|V,\theta)E(H,V) \\
&-\sum_H Q_C(H|V,\theta)\ln Q_C(H|V,\theta). \quad (12)
\end{aligned}
$$

By careful choice of the $Q_C$ distribution, we can arrange for the summations over $H$ in the first two terms to be tractable.

The use of a lower bound for learning is particularly attractive since if we adjust the parameters so as to increase the lower bound this must increase the true log likelihood and/or modify the true conditional distribution to be closer to the approximating distribution (in the sense of KL divergence) thereby making the approximation more accurate. This can be interpreted as a generalized E-step in an EM (expectation-maximization) algorithm (Neal and Hinton, 1998) in which the subsequent optimization of the model parameters corresponds to the M-step. If we allowed arbitrary distributions $Q_C$ instead of restricting attention to a parametric family, we would recover the conventional E-step of the standard EM algorithm (Dempster *et al.*, 1977).

Unfortunately, the term $-\ln Z$ involving the partition function involves summing over exponentially many configurations of the variables and hence remains intractable. We therefore apply the variational framework to this term also by introducing an approximating distribution $Q_F(H,V|\phi)$ over the joint space, where $\phi$ denotes a vector of parameters. In this case we obtain an upper bound on $-\ln Z$ of the form

$$
\begin{aligned}
-\ln Z &= -\ln\left\{\sum_H \sum_V \exp(-E(H,V))\right\} \\
&\leq \sum_H \sum_V Q_F(H,V|\phi)E(H,V) \\
&\quad + \sum_H \sum_V Q_F(H,V|\phi)\ln Q_F(H,V|\phi) \\
&\triangleq \mathcal{L}_F. \qquad\qquad (13)
\end{aligned}
$$

However, the combination of upper and lower bound is not itself a bound. The absence of a rigorous bound is a consequence of the use of an undirected graph, since the $\ln Z$ term does not arise in the case of directed graphs (Bayesian networks).

## 2.1 Mean Field Theory

Mean field theory for Boltzmann machines (Peterson and Anderson, 1987; Hinton, 1989) can be formulated within the variational framework by choosing variational distributions $Q$ which are completely factorized over the corresponding variables. The most general factorized distribution is obtained by allowing each marginal distribution to be governed by its own independent mean field parameter, which we denote by $\boldsymbol{\mu} = \{\mu_i\}$ in the case of the conditional distribution, and $\mathbf{m} = \{m_i\}$ in the case of the joint distribution. Thus we consider

$$
Q_C(H|V,\boldsymbol{\mu}) = \prod_{i \in H} \left(\frac{1+\mu_i}{2}\right)^{\frac{1+h_i}{2}} \left(\frac{1-\mu_i}{2}\right)^{\frac{1-h_i}{2}}
$$

$$
(14)
$$

$$
Q_F(H,V|\mathbf{m}) = \prod_{i \in H,V} \left(\frac{1+m_i}{2}\right)^{\frac{1+s_i}{2}} \left(\frac{1-m_i}{2}\right)^{\frac{1-s_i}{2}}
$$

$$
(15)
$$

Using (2), (12) and (13) we then obtain the following approximation to the log likelihood

$$
\begin{aligned}
\mathcal{L}_{\text{mft}} &= \mathcal{L}_{\text{C}} + \mathcal{L}_{\text{F}} \\
&= \sum_i \sum_{j>i} w_{ij}\mu_i\mu_j + \sum_i \mathcal{H}\left(\frac{1+\mu_i}{2}\right) \\
&\quad - \sum_i \sum_{j>i} w_{ij}m_im_j - \sum_i \mathcal{H}\left(\frac{1+m_i}{2}\right) \quad (16)
\end{aligned}
$$

where $\mu_i$ is defined to be equal to the observed value in the case of clamped units. Here we have defined the binary entropy given by

$$
\mathcal{H}(m) = -p\ln p - (1-p)\ln(1-p). \quad (17)
$$

Note how the assumption of a factorized distribution has allowed the summations over the exponentially many terms to be expressed in terms of a polynomial sum.

We can now optimize the mean field parameters by finding the stationary points of (16) with respect to $\mu_i$ and $m_i$, leading to the following fixed point equations

$$
\mu_i = \tanh\left(\sum_j w_{ij}\mu_j\right) \quad (18)
$$

$$
m_i = \tanh\left(\sum_j w_{ij}m_j\right). \quad (19)
$$

which can be solved iteratively.

Once the mean field parameters have been determined we can update the model parameters using gradient-based optimization techniques. This requires evaluation of the derivatives of the objective function, given by

$$
\frac{\partial \mathcal{L}_{\text{mft}}}{\partial w_{ij}} = \mu_i\mu_j - m_im_j. \quad (20)
$$

Thus we see that the gradients have been expressed in terms of simple products of mean field parameters, which themselves can be determined by iterative solution of deterministic equations. The resulting learning algorithm is computationally efficient compared with stochastic optimization of the true log likelihood. Comparison of (20) with (6) shows how the expectations have been replaced with deterministic approximations.

In a practical setting it is often useful to introduce a temperature parameter as in (1). For large values of $T$ the true distribution of the parameters is smoothed out and the variational optimization is simplified. The value of $T$ can then be slowly reduced to $T = 1$ (this is called annealing) while continuing to update the mean field parameters. This helps the variational approximation to find better solutions by avoiding locally optimal, but globally suboptimal, solutions.

## 3  Mixture Representations

We have already noted that mean field theory, while computationally tractable, frequently fails to find satisfactory solutions (Galland, 1993). The origin of the difficulty lies in the rather drastic assumption underlying mean field theory of a fully factorized distribution. One consequence is that mean field theory is only able to approximate uni-modal distributions with any accuracy. In practice we will often expect the true distributions to be multi-modal, particularly in the case of the joint distribution corresponding to the 'free' phase. If, for example, the data set consists of sub-populations, or clusters, then the joint distribution will necessarily be multi-modal. However, it may be the case that each of the conditional distributions can be well approximated by a uni-modal distribution (so that only one hidden 'cause' is required to explain each observation). Indeed, this will trivially be the case for models with no hidden variables. Thus we expect the problems with mean field theory to arise primarily in its approximation to the statistics of the unclamped phase.

We address this difficulty by introducing a variational approximation consisting of a *mixture* of factorized distributions (Jaakkola and Jordan, 1997; Bishop *et al.*, 1997). This is used to approximate the free phase, while standard mean field theory is used for the clamped phase[1]. We therefore consider an approximating distribution of the form

$$
Q_{\text{mix}}(S) = \sum_{l=1}^{L} \alpha_l Q_{\text{F}}(S|l) \quad (21)
$$

where each of the components $Q_{\text{F}}(S|l)$ is a factorized distribution with its own variational parameters

$$
Q_{\text{F}}(S|l) = \prod_{i \in H,V} \left(\frac{1+m_{li}}{2}\right)^{\frac{1+s_i}{2}} \left(\frac{1-m_{li}}{2}\right)^{\frac{1-s_i}{2}}. \quad (22)
$$

The mixing coefficients $\alpha_l$ satisfy $\alpha_l \geq 0$ and $\sum_l \alpha_l = 1$. Using the variational distribution (21) we can express $\mathcal{L}_{\text{F}}$ from (13) in the form (Jaakkola and Jordan, 1997)

$$
\mathcal{L}_{\text{F}}(Q_{\text{mix}}) = \sum_{l=1}^{L} \alpha_l \mathcal{L}_{\text{F}}(Q_{\text{F}}(S|l)) + I(l, S) \quad (23)
$$

---

[1]It is straightforward to extend the procedure to use mixture distributions for the clamped phase also, if this is thought necessary in some particular application.

where $I(l, S)$ is the mutual information between the component label $l$ and the variables $S$ given by

$$I(l, S) = \sum_l \sum_S \alpha_l Q_F(S|l) \ln \left\{ \frac{Q_F(S|l)}{Q_{\text{mix}}(S)} \right\}. \quad (24)$$

The first term is simply a linear combination of mean field contributions, and as such it provides no improvement over the simple mean field bound (since the optimal bound would be obtained by setting all of the $\alpha_l$ to zero except for the one corresponding to the $Q_F(S|l)$ giving the tightest bound, thereby recovering standard mean field theory). It is the second, mutual information, term which allows the mixture representation to give an improved relative to mean field theory. However, the mutual information again involves an intractable summation over the states of the variables. In order to be able to treat it efficiently we first introduce a set of 'smoothing' distributions $R(S|l)$, and rewrite the mutual information (24) in the form

$$I(l, S) = \sum_{l,S} \alpha_l Q_F(S|l) \ln R(S|l) - \sum_l \alpha_l \ln \alpha_l$$
$$- \sum_{l,S} \alpha_l Q_F(S|l) \ln \left\{ \frac{R(S|l)}{\alpha_l} \frac{Q_{\text{mix}}(S)}{Q_F(S|l)} \right\}. \quad (25)$$

It is easily verified that (25) is equivalent to (24) for arbitrary $R(S|l)$. We next make use of the following inequality

$$- \ln x \geq - \lambda x + \ln \lambda + 1 \quad (26)$$

to replace the logarithm in the third term in (25) with a linear function (conditionally on the component label $l$). This yields a lower bound on the mutual information given by $I(l, S) \geq I_\lambda(l, S)$ where

$$\begin{aligned} I_\lambda(l, S) &= \sum_l \sum_S \alpha_l Q(S|l) \ln R(S|l) - \sum_l \alpha_l \ln \alpha_l \\ &\quad - \sum_l \lambda_l \sum_S R(S|l) Q_{\text{mix}}(S) \\ &\quad + \sum_l \alpha_l \ln \lambda_l + 1. \end{aligned} \quad (27)$$

The summations over configurations $S$ in (27) can be performed analytically if we assume that the smoothing distributions $R(S|l)$ factorize.

In order to obtain the tightest bound within the class of approximating distributions, we can maximize the bound with respect to the variational parameters $m_{li}$, the mixing coefficients $\alpha_l$, the smoothing distributions $R(S|l)$ and the variational parameters $\lambda_l$. This yields straightforward re-estimation equations, for which the details can be found in Jaakkola and Jordan (1997).

Once the variational approximations to the joint and conditional distributions have been optimized, the derivatives of the cost function are evaluated using

$$\frac{\partial \mathcal{L}_{\text{mix}}}{\partial w_{ij}} = \mu_i \mu_j - \sum_l \alpha_l m_{li} m_{lj}. \quad (28)$$

These derivatives are then used to update the weights using a gradient-based optimization algorithm. The learning algorithm then alternates between optimization of the variational approximation (analogous to an E-step) and optimization of the weights (analogous to an M-step).

## 4   Results: Inference

Our variational framework allows expectations of the form $\langle s_i s_j \rangle$ to be approximated by deterministic expressions involving variational parameters, of the form $\sum_l \alpha_l m_{li} m_{lj}$, in which standard mean field theory corresponds to the case of just one component in the mixture. We now investigate how well this approach is able to approximate the true expectations, and how the approximation improves as the number of components in the mixture is increased.

For this purpose we consider small networks such that the (exponentially large) summation over states can be performed exactly, thereby allowing us to compare the variational approximation to the true expectation. The networks have ten variables and are fully interconnected, and hence have 55 independent parameters including biases. None of the units are clamped. Evaluation of the expectations involves summing over $2^{10} = 1024$ configurations. We have generated 100 networks at random in which the weights and biases have been chosen from a uniform distribution over $(-1, 1)$. For each network we approximate the joint distribution of variables using mixture distributions involving $L$ components, where $L = 1, \ldots, 10$, and the temperature parameter $T$ was annealed in 8 steps from $T = 60$ to $T = 1$. In Figure 1 we show plots of the histograms of the differences between the approximate and exact expectations, given by

$$\sum_{l=1}^L \alpha_l m_{li} m_{lj} - \langle s_i s_j \rangle, \quad (29)$$

together with a summary of the behaviour of the sum-of-squares of the differences (summed over all 100 networks) versus the number $L$ of components in the mixture. We see that there is a clear and systematic improvement in the accuracy with which the expectations are approximated as the number of components in the mixture is increased.

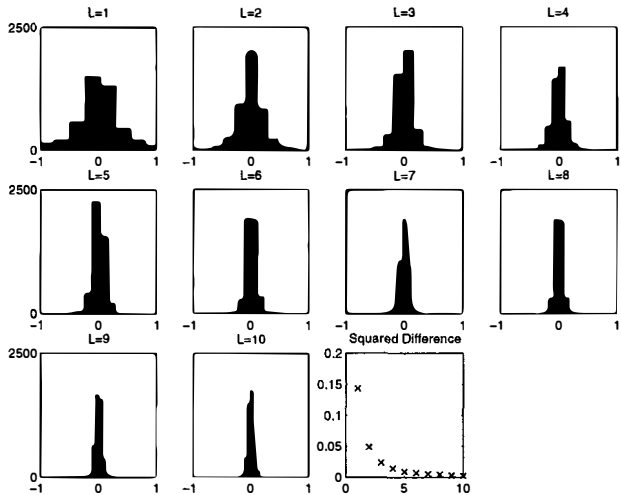

Figure 1: Histograms of the differences between true and approximate expectations for 100 randomly generated networks each having 55 independent parameters, for different numbers $L$ of components in the mixture approximation, together with a summary of the dependence of the sum-of-squares of the differences on $L$.

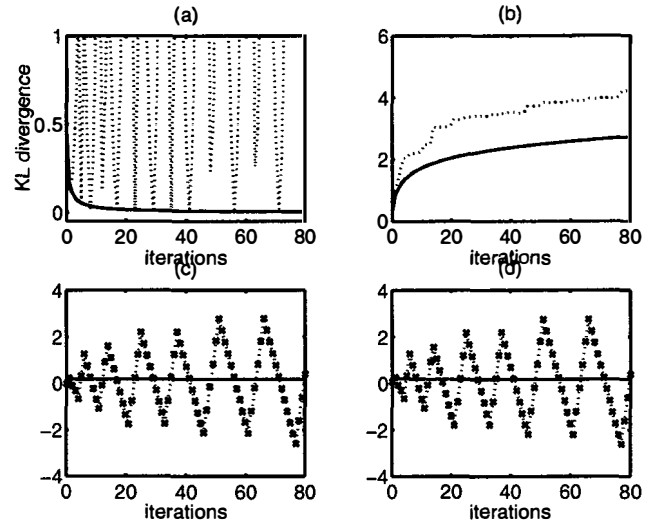

Figure 2: Results from learning in the toy problem. (a) The KL divergence between the true distribution of the training data and the distribution represented by the network as a function of the number of iterations of learning. The results from mean field theory (dotted curve) are wildly unstable whereas the results obtained using a mixture model (dashed curve) are well behaved and almost indistinguishable from those found using the true log likelihood (solid curve). The remaining figures show the corresponding evolution of the weight parameter (b) and the two bias parameters (c) and (d) for the three approaches.

## 5 Results: Learning

In the previous section we have seen how the use of mixture representations can lead to improved accuracy of inference compared with standard mean field theory. We now investigate the extent to which improved inference leads to improved learning. For simplicity we use simple gradient ascent learning, with gradients evaluated using (20) or (28). In Section 5.1 we consider a simple toy problem designed to have a multi-modal unconditional distribution, and then in Section 5.2 we apply our approach to a problem involving images of hand-written digits.

### 5.1 Toy Problem

As a simple example of a problem leading to a multi-modal distribution we follow Kappen and Rodriguez (1998) and consider a network consisting of just two visible nodes, together with a data set consisting of two copies of the pattern $(1, 1)$ and one copy of the pattern $(-1, -1)$. In this case the distribution in the unclamped phase needs to be bimodal for the network to have learned a solution to the problem. Due to the small size of the network, comparison with exact results is straightforward. We apply standard mean field theory, and compare it with a mixture model having two components, and with learning using the exact log likelihood gradient. In the inference stage, the variational parameters are iteratively updated until the cost function $\mathcal{L}_F$ changes by no more than 0.01% (up to a maximum of 20 iterations). No annealing was used. The network is initialized using parameters drawn from a zero-mean Gaussian distribution having a standard deviation of 0.1, and learning is by gradient ascent with a learning rate parameter of 0.25. The results are shown in Figure 2.

Mean field theory seen to be quite unstable during learning. In particular, the bias parameters undergo systematic oscillations. To investigate this further we plot an expanded region of the training curve from Figure 2 (c) in Figure 3 together with the mean field parameters at each learning step. We see that the unimodal approximating distribution of mean field theory is oscillating between the two potential modes of the joint distribution, as the algorithm tries to solve this multi-modal problem.

This phenomenon can be analysed in terms of the stability structure of the mean field solutions. We find that, for the first few iterations of the learning algorithm when the weight value is small, the mean field equations (18) exhibit a single, stable solution with small values of the mean field parameters. However, once the weight value grows beyond a critical value, two stable solutions (and one unstable solution) appear whose values depend on the bias parameters. Evolution of the bias parameters modifies the shape of the stability diagram and causes the mean field so-lution to oscillate between the two stable solutions, as the parameter vector repeatedly 'falls off' the edges of the cusp bifurcation (Parisi, 1988).

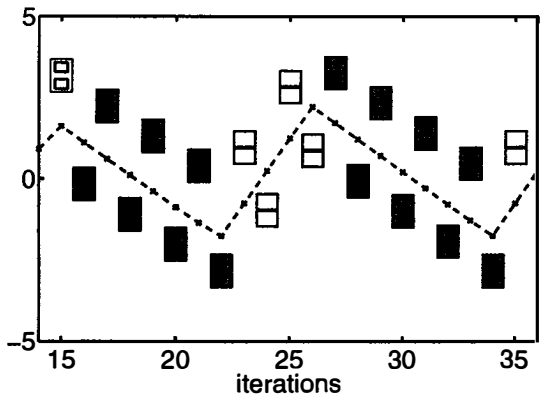

Figure 3: Expanded plot of the evolution of one of the bias parameters during training with mean field theory. Also shown are plots of the mean field parameters, represented as 'Hinton' diagrams in which each of the two parameters is denoted by a square whose area is proportional to the parameter value and white denotes negative values, while grey denotes positive values.

## 5.2  Handwritten Digits

As a second example of learning we turn to a more realistic problem involving hand-written digits[2] which have been pre-processed to give 8 × 8 binary images. We extracted a data set consisting of 700 examples of each of the digits 0 through 9. Examples of the training data are shown in Figure 4.

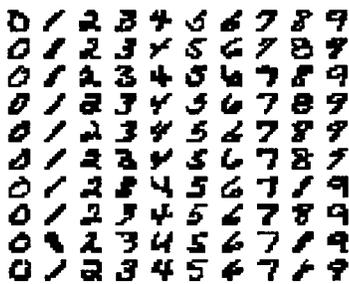

Figure 4: Examples of the hand-written digits from the training set.

The networks consisted of 64 visible nodes in an 8 × 8 grid, with each visible node connected to its neighbours on both diagonals and in the horizontal and vertical directions. The network also had ten hidden nodes which are fully connected with each other and with all the visible nodes. Additionally all nodes had an associated bias. An annealing schedule was used during the inference steps involving 7 successive values of the temperature parameter going from $T = 100$ down to $T = 1$. Learning was achieved through 30 iterations of gradient ascent in the parameters $w_{ij}$, with a learning rate of $0.1/N$ where $N = 7,000$ is the total number of patterns in the training set.

[2]Available on the CEDAR CDROM from the U.S. Postal Service Office of Advanced Technology.

Due to the size of the network it is no longer possible to perform exact calculations for the unclamped distribution. We therefore compare standard mean field theory with a mixture distribution having ten components. Figure 5 shows the evolution of the cost functions $\mathcal{L}_{\mathrm{mft}}$ and $\mathcal{L}_{\mathrm{mix}}$. Again we see that mean field

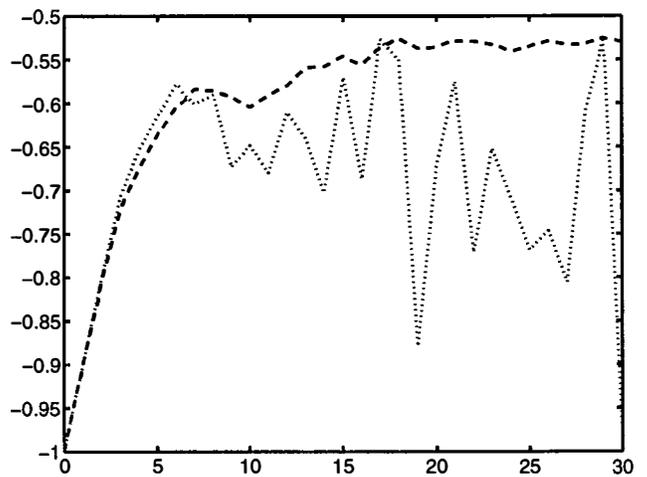

Figure 5: Evolution of the cost functions $\mathcal{L}_{\mathrm{mft}}$ (dotted curve) and $\mathcal{L}_{\mathrm{mix}}$ (dashed curve) for the digits problem.

theory is relatively unstable compared to the mixture model.

Figure 6 shows the evolution of the variational parameters from the unclamped phase (plotted for the visible units only), as this provides insight into the behaviour of the algorithm. We see that simple mean field theory exhibits substantial 'mode hopping', while the components of the mixture distribution are much more stable (although some tendency to mode hop is still evident, suggesting that a larger number of components in the distribution may be desirable).

## 6  Discussion

In this paper we have shown how the fundamental limitations of mean field theory for Boltzmann machines can be overcome by using variational inference based on mixture distributions. Preliminary results indicate a significant improvement over standard mean field theory for problems in which the joint distribution over visible and hidden units is multi-modal.

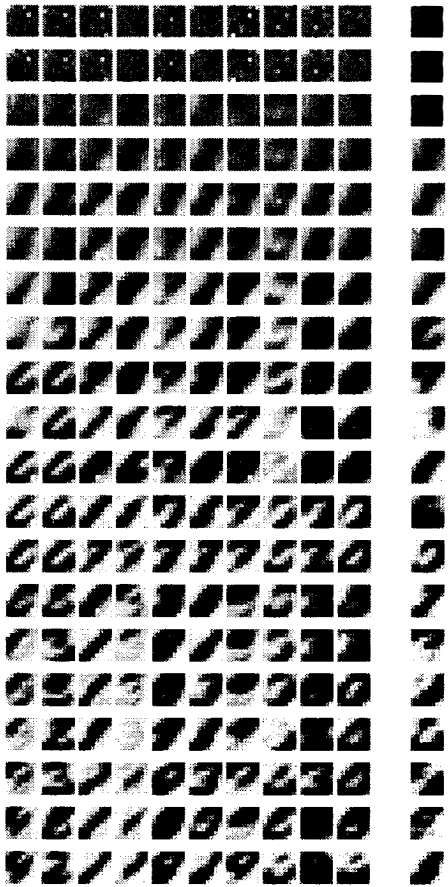

Figure 6: Variational parameters from $Q_F(H, V)$ in which successive rows correspond to successive iterations of learning running from 1 to 20. The right most column corresponds to mean field theory while the first ten columns correspond to the ten components in the mixture model. In each case only the parameters corresponding to the visible variables are shown.

Although the use of mixtures is somewhat more costly computationally than standard mean field theory (scaling roughly linearly in the number of components in the mixture) it should be remembered that the optimization of the corresponding $Q$ distribution has to be done only once for each pass through the data set, while a separate optimization has to be done for each clamped distribution corresponding to every data point. For moderate to large data sets, the overall increase in computational cost compared with standard mean field theory will therefore be negligible. One consequence it that it is possible to run this algorithm with a very large number of components in the mixture distribution while still incurring little computational penalty compared with standard mean field theory.

Our experimental results have also revealed an interesting phenomenon whereby the uni-modal distribution of mean field theory appears to oscillate between modes in the joint distribution during learning.

## Acknowledgements

We are grateful to Tommi Jaakkola for helpful comments. Also, we would like to thank David MacKay for a stimulating discussion about mean field theory.

## References

Ackley, D., G. Hinton, and T. Sejnowski (1985). A learning algorithm for Boltzmann machines. *Cognitive Science* **9**, 147–169.

Bishop, C. M., T. Jaakkola, M. I. Jordan, and N. Lawrence (1997). Approximating posterior distributions in belief networks using mixtures. Technical report, Neural Computing Research Group, Aston University, Birmingham, UK. To appear in Proceedings NIPS'97.

Dempster, A. P., N. M. Laird, and D. B. Rubin (1977). Maximum likelihood from incomplete data via the EM algorithm. *Journal of the Royal Statistical Society, B* **39** (1), 1–38.

Galland, C. C. (1993). The limitations of deterministic Boltzmann machines. *Network: Computation in Neural Systems* **4** (3), 355–379.

Hinton, G. (1989). Deterministic boltzmann machine performs steepest descent in weight-space. *Neural Computation* **1** (1), 143–150.

Jaakkola, T. and M. I. Jordan (1997). Approximating posteriors via mixture models. To appear in Proceedings NATO ASI *Learning in Graphical Models*, Ed. M. I. Jordan. Kluwer.

Jordan, M. I., Z. Gharamani, T. S. Jaakkola, and L. K. Saul (1998). An introduction to variational methods for graphical models. In M. I. Jordan (Ed.), *Learning in Graphical Models*. Kluwer.

Kappen, H. J. and F. B. Rodriguez (1998). Efficient learning in Boltzmann machines using linear response theory. Technical report, Department of Biophysics, University of Nijmegen.

Neal, R. M. and G. E. Hinton (1998). A new view of the EM algorithm that justifies incremental and other variants. In M. I. Jordan (Ed.), *Learning in Graphical Models*. Kluwer.

Parisi, G. (1988). *Statistical Field Theory*. Redwood City, CA: Addison-Wesley.

Peterson, C. and J. R. Anderson (1987). A mean field learning algorithm for neural networks. *Complex Systems* **1**, 995–1019.